\documentclass[letterpaper]{article} 
\usepackage{aaai25}  
\usepackage{times}  
\usepackage{helvet}  
\usepackage{courier}  
\usepackage[hyphens]{url}  
\usepackage{graphicx} 
\usepackage{natbib}  
\usepackage{caption} 
\usepackage{algorithm}
\usepackage{algorithmic}

\usepackage{newfloat}
\usepackage{listings}
\DeclareCaptionStyle{ruled}{labelfont=normalfont,labelsep=colon,strut=off} 
\floatstyle{ruled}
\newfloat{listing}{tb}{lst}{}
\floatname{listing}{Listing}
\title{Think Outside the Data: Colonial Biases and Systemic Issues in Automated Moderation Pipelines for Low-Resource Languages}
\author {
    Farhana Shahid\textsuperscript{\rm 1},
    Mona Elswah\textsuperscript{\rm 2, 3},
    Aditya Vashistha\textsuperscript{\rm 1}
}
\affiliations {
    \textsuperscript{\rm 1}Cornell University\\
    \textsuperscript{\rm 2}Center for Democracy and Technology and
    \textsuperscript{\rm 3}University of Exeter\\
    fs468@cornell.edu, m.a.i.elswah@exeter.ac.uk, adityav@cornell.edu
}

\newcommand{\parabold}[1]{\noindent\textbf{#1.}}

\begin{document}

\maketitle

\begin{abstract}
Most social media users come from the Global South, where harmful content usually appears in local languages. Yet, AI-driven moderation systems struggle with low-resource languages spoken in these regions. Through semi-structured interviews with 22 AI experts working on harmful content detection in four low-resource languages: Tamil (South Asia), Swahili (East Africa), Maghrebi Arabic (North Africa), and Quechua (South America)--we examine systemic issues in building automated moderation tools for these languages. Our findings reveal that beyond data scarcity, socio-political factors such as tech companies' monopoly on user data and lack of investment in moderation for low-profit Global South markets exacerbate historic inequities. Even if more data were available, the English-centric and data-intensive design of language models and preprocessing techniques overlooks the need to design for morphologically complex, linguistically diverse, and code-mixed languages. We argue these limitations are not just technical gaps caused by ``data scarcity'' but reflect structural inequities, rooted in colonial suppression of non-Western languages. We discuss multi-stakeholder approaches to strengthen local research capacity, democratize data access, and support language-aware solutions to improve automated moderation for low-resource languages. 

\end{abstract}

%

\section{Introduction}

The largest and fastest-growing social media user base comes from the Global South, where billions generate content in their local languages. This growth has fueled misinformation, hate speech, and incitement to violence in non-English languages, contributing to serious human rights violations~\cite{Samuels-2020, Milmo-2021, Yibeltal-2023}. However, the moderation systems deployed by tech companies prioritize English-speaking users in the West~\cite{Legon-2020, Popli-2021}, leaving harmful content in Global South languages largely unchecked, increasing social harm and political divides~\cite{Nigatu-2024-YT, Samuels-2020, Milmo-2021}. Moreover, flawed moderation systems often censor benign content in non-English languages and silence marginalized voices~\cite{Elswah-arabic-2024}.

Regardless of the content modality, tech companies heavily rely on text-based methods, such as analyzing captions, transcripts, and subtitles, to automatically moderate harmful content~\cite{audio-2025, video-2025}. However, this approach poses challenges for many languages spoken in the Global South, which are historically considered \textit{``low-resourced''} due to the lack of high-quality datasets needed to train AI models~\cite{Rowe-2022, nicholas-2023, nigatu-2024}. However, data scarcity tells only part of the story. Economic and political oppression, insufficient human expertise, and limited access to digital infrastructures further exacerbate the \textit{``low-resourcedness''} of these languages~\cite{nigatu-2024}. Moreover, framing the problem solely as one of data scarcity overlooks broader challenges across the moderation pipeline, such as annotation, model training, and deployment. 
To address this critical gap, we examine the systemic barriers hindering equitable moderation for low-resource languages and explore actionable pathways to improve these systems. Specifically, we ask:

\begin{itemize}
    \item[\textbf{RQ1:}] What systemic barriers impact automated moderation pipelines for low-resource languages? 
    \item[\textbf{RQ2:}] How might we improve automated moderation for low-resource languages? 
\end{itemize}

To address these questions, we conducted semi-structured interviews with 22 AI researchers and practitioners, specializing in harmful content detection and developing automated tools for diverse low-resource languages, such as Tamil from South Asia, Swahili in East Africa, Maghrebi Arabic from North Africa, and Quechua in South America. 

Our findings reveal a spectrum of systemic issues beyond data scarcity impacting the automated moderation pipeline for low-resource languages. Many participants criticized tech companies' data restriction policy for hindering moderation research in the Global South. They pointed out that company's use of biased machine translation systems, Western-centric toxicity models, and poor language detection tools---overlook the cultural nuances of online harms and language evolution in the Global South. They emphasized that even with more data, current English-centric design of preprocessing techniques (e.g., tokenization, stemming) and language models disregard the linguistic diversity, morphological complexity, and dynamic evolution of languages through code-mixing and code-switching, which are often absent in English. For instance, unlike English which has a relatively fixed word order~\cite{bender-2009}, Tamil, Swahili, Arabic, and Quechua have agglutinative property, meaning they can form thousands of complex words from a single root. Data-driven models primarily trained on English typically fail to infer these linguistic properties that do not exist in English. As a result, words that frequently appear in sexual harassment, such as Tamil word \textit{Mualichhu} (meaning, n**ples) incorrectly gets stemmed to \textit{Mulai-} (meaning, sprout) and goes undetected by models. 

Drawing on these findings, we use coloniality as a lens to critically examine how tech companies perpetuate digital colonialism~\cite{kwet-2019}, prioritizing profit over user safety in less profitable Global South markets~\cite{nicholas-2023}. These companies not only monopolize data extracted from the next billion users~\cite{coleman-2018, couldry-2019}, but also rely on biased sources for moderation—reinforcing harmful narratives about formerly colonized populations. We highlight how the English-centric, one-size-fits-all design of moderation tools reflects a colonial impulse by ignoring the linguistic diversity of Global South languages. Improving moderation for these languages requires more than technical fixes, as competing stakeholder priorities demand deeper systemic change. The key contributions of our work are as follows:

\begin{itemize}
    \item A qualitative study using coloniality as a lens to examine how historical power imbalances skew automated moderation for low-resource languages in the Global South.
    \item A guideline to improve moderation for low-resource languages, acknowledging the practical and systemic issues.
\end{itemize}
\section{Related Work}

\subsection{Content Moderation in the Global South}
Content moderation refers to reviewing user-generated content to see if it aligns with tech company's policies on what should be allowed on the platform. Most companies use a mix of manual human reviews and automated AI models~\cite{gorwa-2019}. However, they often lack financial incentives to invest in moderation resources for 
less profitable markets in the Global South~\cite{deGregorio-2023, nicholas-2023}. 
For instance, Meta funnels 87\% of its global misinformation budget to the United States (US), despite Americans comprising only 10\% of its user base~\cite{Popli-2021}. The disparity is more glaring when companies swiftly handle harmful content from European countries that either offer strong economic incentives or hold geopolitical interest for the US (e.g., Russia-Ukraine war)~\cite{deGregorio-2023}. In contrast, tech companies have been less proactive in countering harmful content festering in many Global South countries~\cite{Milmo-2021, wong-Ernst-2021, wong-harding-2021, Yibeltal-2023}, while unjustly removing culturally and politically legitimate content from this region~\cite{Elswah-arabic-2024, Shahid-2023}.

Prior research links poor moderation in the Global South to tech companies' reliance on automated systems primarily trained on English and the lack of data in low-resource languages~\cite{nicholas-2023, deGregorio-2023, nigatu-2024}. In contrast, little focus is given to other critical stages of moderation pipelines, such as who annotates what is harmful or what assumptions are made about deploying these models in complex, low-resource environments. To fill this critical gap, we explore the systemic challenges AI experts face throughout the moderation pipeline, placing these issues in the context of historical and colonial marginalization of non-English languages. We now present scholarly work critically examining power and control in content moderation systems. 

\subsection{Coloniality in Content Moderation}
Decolonial scholars argue that colonial power structures persist by exploiting the labor and resources of formerly colonized populations while maintaining Western dominance over governance and knowledge production~\cite{quijano-2000, quijano-2007}. Decolonial computing challenges who designs technology, where it's developed, and how it impacts marginalized communities~\cite{ali2016brief}. Whereas, postcolonial scholars critique the imposition of Western-centric technologies on diverse cultural contexts~\cite{Irani-2010}. Scholars have applied these lenses to examine content moderation systems.

\citet{Shahid-2023} used decoloniality to show how tech companies often impose Western values as global community standards, ignoring local socio-cultural norms when assessing online harms in the Global South. They draw parallels between Western-centrism in community guidelines and colonial suppression of Indigenous and marginalized communities' diverse ways of being, imposing Euro-modern rationality as the only legitimate way~\cite{said-2000, gramsci-2020, quijano-2007}. Similarly, \citet{siapera-2022} offers a decolonial critique of how tech companies dismiss input from racialized users when shaping policies against racist hate speech. She argues the race-blind, \textit{``neutral''} moderation technologies mirror colonial legacies, criminalizing the identities and lived experiences of racialized communities as inferior~\cite{quijano-2007questioning, benjamin2023race}.  

In addition, scholars highlight how tech companies build AI moderation systems exploiting the labor and trauma of low-wage moderators from the Global South~\cite{siapera-2022, Shahid-2023, Elswah-2024}. Tech companies often treat these moderators as disposable, concealing the nature of the work during recruitment, and evading responsibility for the harms moderators face~\cite{ahmad-2022, Elswah-2024}. 

Moreover, moderation errors disproportionately harm marginalized communities, whose voices have long been silenced. AI models often reinforce systemic racism and heteronormative patriarchy by mislabeling Black and queer vernacular as toxic~\cite{bhattacharyya-2018, sap-2019, mohamed-2020}. Shaped by Western norms, these systems frequently misclassify culturally appropriate content in non-Western contexts as harmful while overlooking actual harmful content~\cite{Shahid-2023}. For instance, Google’s Perspective API underestimates toxicity in Swahili and Hindi but rates similar content in English and German more accurately~\cite{udupa-2023}. \citet{udupa-2023} argues these biases persist because moderation systems inherit Eurocentric, colonial frameworks that justify unequal resource allocation across languages and regions.

We contribute to existing literature by critically examining the often-overlooked socio-political dynamics embedded across the automated moderation pipeline—from data collection to labeling, training, and evaluation. Since many studies overlook historical contexts when discussing power asymmetries in AI~\cite{Ovalle-2023}, we address this gap by interrogating the normative assumptions and design paradigms behind AI-driven moderation through the lenses of power and control. Drawing on interviews with AI experts, we explore how current language technologies shape moderation in diverse Global South contexts (RQ1) and how these practices might be improved (RQ2).

\section{Methodology}

To examine disparities in automated moderation pipelines, we interviewed 22 AI experts specializing in diverse low-resource languages spoken across the Global South. 

\parabold{Low-Resource Languages} We selected four diverse languages from different parts of the Global South. These are: Tamil (South Asia), Swahili (East Africa), Maghrebi Arabic (North Africa), and Quechua (South America) (see Table~\ref{tab:language-feature}). All these languages are considered low-resourced, despite having millions of speakers. UNESCO even declared Quechua as a vulnerable language due to systemic discrimination against Indigenous Quechua speakers in South America~\cite{bank-2014}. Due to limited resources, moderation errors are typically high for these languages. For instance, tech companies have repeatedly failed to address ethnic hate speech in Swahili~\cite{witness-2022} and harmful content in Arabic~\cite{Elswah-arabic-2024}, while unjustly removing Tamil news articles as dangerous speech~\cite{Biddle-2022} and shadowbanning Arabic content on Palestine~\cite{Elswah-arabic-2024}. 

\begin{table*}[ht]
\caption{Various characteristics of four low-resource languages featured in this study.}
\label{tab:language-feature}
\centering
\begin{tabular}{l|l|l|l|l|}
\cline{2-5}
                                         & \multicolumn{1}{c|}{\textbf{Tamil}}                                                        & \multicolumn{1}{c|}{\textbf{Swahili}}                                               & \multicolumn{1}{c|}{\textbf{Maghrebi Arabic}}                                     & \multicolumn{1}{c|}{\textbf{Quechua}}                                               \\ \hline
\multicolumn{1}{|l|}{Number of speakers} & 80 million                                                                                 & 100 million                                                                         & 88 million                                                                        & 8 million                                                                           \\ \hline
\multicolumn{1}{|l|}{Geographic region}  & \begin{tabular}[c]{@{}l@{}}South Asia: Tamil Nadu \\ (India), Sri Lanka, etc.\end{tabular} & \begin{tabular}[c]{@{}l@{}}East Africa: Kenya, \\ Tanzania, etc.\end{tabular}       & \begin{tabular}[c]{@{}l@{}}North Africa: Morocco,\\ Algeria, Tunisia, etc.\end{tabular} & \begin{tabular}[c]{@{}l@{}}Andes: Bolivia,\\ Peru, Ecuador, etc.\end{tabular}       \\ \hline
\multicolumn{1}{|l|}{Language family}    & Dravidian                                                                                  & Bantu                                                                 & Semitic                                                            & \begin{tabular}[c]{@{}l@{}}Quechuan \end{tabular}                    \\ \hline
\multicolumn{1}{|l|}{Grammar}            & \begin{tabular}[c]{@{}l@{}}Agglutinative, subject-\\ object-verb (SOV)\end{tabular}        & \begin{tabular}[c]{@{}l@{}}Agglutinative, subject-\\ verb-object (SVO)\end{tabular} & \begin{tabular}[c]{@{}l@{}}Root based, verb-\\ subject-object (VSO)\end{tabular}  & \begin{tabular}[c]{@{}l@{}}Agglutinative, subject-\\ object-verb (SOV)\end{tabular} \\ \hline
\multicolumn{1}{|l|}{Colonial influence} & British                                                                                    & \begin{tabular}[c]{@{}l@{}}Portugese, German, \\ British, Arabic\end{tabular}       & French, Spanish, Italian                                                          & Spanish                                                                             \\ \hline
\end{tabular}
\end{table*}

\parabold{Participants} We recruited people, who either (1) worked on automatic detection of harmful content, or (2) developed language models and tools in Tamil, Swahili, Maghrebi Arabic, or Quechua. We used purposive and snowball sampling to recruit 22 participants (see Table~\ref{tab:demographic}). Among them, six specialized in Tamil, six in Swahili, five in Maghrebi Arabic, and three in Quechua. Most of them (n=15) were native speakers of one of these languages. Many of our participants were affiliated with academia (n=12) and trust and safety teams at Meta, OpenAI, and TikTok (n=4). Some worked for trust and safety vendors, who built moderation tools and datasets for different clients (n=2) and local AI startups (n=5). Some participants held multiple roles. Five self-identified as women and the rest as men. All participants had experience living in the Global South, such as Kenya, Tanzania, India, Sri Lanka, Peru, Morocco, and Egypt. Half of them were affiliated with Western institutions and were based in North America and Europe during the interview. 

\begin{table*}[ht]
\caption{Demographics of participants in our study.}
\label{tab:demographic}
\centering
\begin{tabular}{|l|c|c|l|c|c|}
\hline
\textbf{ID} & \textbf{\begin{tabular}[c]{@{}c@{}}Language\\ expertise\end{tabular}} & \textbf{Role}                                                         & \textbf{ID} & \textbf{\begin{tabular}[c]{@{}c@{}}Language\\ expertise\end{tabular}} & \textbf{Role}                 \\ \hline
P1          & Tamil                                                                 & \begin{tabular}[c]{@{}c@{}}Professor, Startup\\ founder in India\end{tabular} & P12         & Swahili                                                               & Professor in Kenya                    \\ \hline
P2          & Tamil                                                                 & Professor in India                                                            & P13         & Swahili                                                               & Master's student in Kenya             \\ \hline
P3          & Tamil                                                                 & PhD Student in Europe                                                         & P14         & Swahili                                                               & Professor in Tanzania                 \\ \hline
P4          & Tamil                                                                 & Master's student in US                                                        & P15         & Swahili                                                               & Startup founder in UK                 \\ \hline
P5          & Tamil                                                                 & Software engineer in India                                                    & P16         & Swahili                                                               & ML engineer in Kenya                  \\ \hline
P6          & Tamil                                                                 & Lecturer in Sri Lanka                                                         & P17         & Swahili                                                               & Industry practitioner at Meta        \\ \hline
P7          & Indic languages                                                       & Trust and safety vendor                                                       & P18         & Arabic                                                                & Industry practitioner at TikTok       \\ \hline
P8          & Indic languages                                                       & Trust and safety vendor                                                       & P19         & Arabic                                                                & Industry practitioner at Meta, OpenAI \\ \hline
P9          & Quechua                                                               & Linguist in Europe                                                            & P20         & Arabic                                                                & Industry practitioner at Meta         \\ \hline
P10         & Quechua                                                               & PhD student in Europe                                                          & P21         & Arabic                                                                & PhD student in UK                     \\ \hline
P11         & Quechua                                                               & Lecturer in Peru                                                              & P22         & Arabic                                                                & Startup founder in Europe            \\ \hline
\end{tabular}
\end{table*}

\parabold{Data Collection and Analysis} We conducted 40-60 minutes long semi-structured interviews with the participants via Zoom. The semi-structured interviews focused on data collection, annotation, preprocessing, and model development in low-resource languages. We asked about the reliability and performance of models and tools they used to detect harmful content. The participants also reflected on biases and challenges they encounter throughout the process and discussed ways to address them. After each interview, we iteratively refined our interview protocol, stopping when the responses reached saturation. After obtaining ethical approvals from IRB, we conducted the interviews in English and audio recorded with the consent of participants. We compensated the participants with \$100 Visa gift cards. 

We transcribed the interviews, performed iterative open coding following reflexive thematic analysis~\cite{braun-2006}, and continuously refined the emerging themes. Our coding process resulted in 441 codes, iteratively merged into 23 subthemes (e.g., model performance, annotation challenges)--which we mapped into different stages of automated moderation pipelines. 

\section{Findings}

\begin{figure*}
    \centering
    \includegraphics[width=\linewidth, trim={0.7cm 0.3cm 0.7cm 0.5cm},clip]{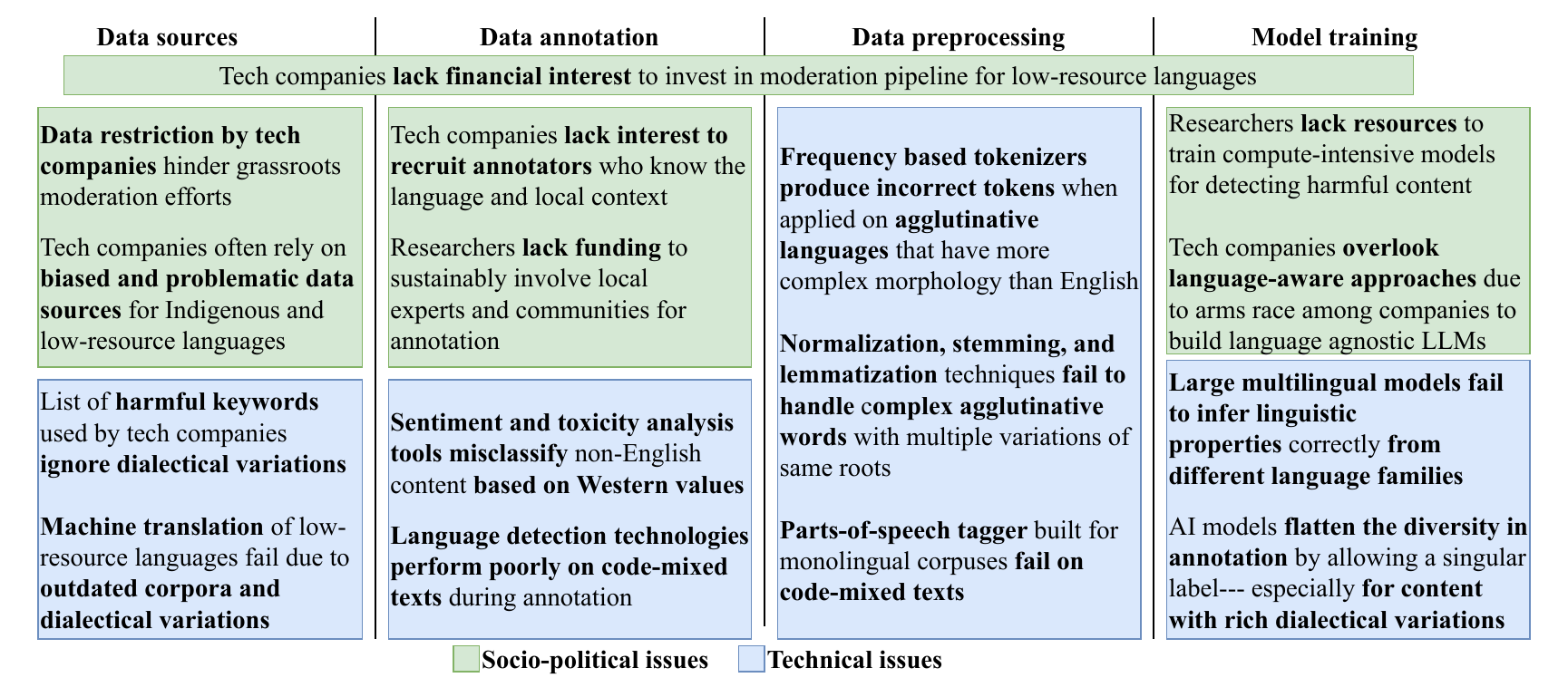}
    \caption{Issues affecting different stages of automated moderation pipeline for low-resource languages.}
    \label{fig:enter-label}
\end{figure*}

In this section, we outline systemic issues in moderating content in low-resource languages throughout automated moderation pipeline: data curation (\ref{dataset}), annotation (\ref{annotation}), preprocessing (\ref{preprocessing}), and model training (\ref{models}).

\subsection{Barriers to Access Data on Harmful Content}\label{dataset}
To detect harmful content, most participants relied on large volumes of user-generated data from platforms like Facebook, X, YouTube, and Reddit. Industry practitioners reported easy access to such data within their companies, while academic researchers pointed to structural barriers, including lack of public datasets in low-resource languages and restricted access to social media data. For instance, during 2018–19, Twitter’s free API limited researchers access to data older than two weeks. P14, a Swahili-focused academic researcher, commented:
\begin{quote}
    \textit{``People frequently used the word `madoadoa' [spots] to spew hatred and violence during the 2007-08 Kenyan election. But that changed in the 2022 election. Bad actors appropriated the popular song `sipangwi' [I am not told what to do] and its plural form `hatupagwingwi' to spread hatred. Unfortunately we neither have access to recent nor past data to study how hate speech tactics have evolved.''}
\end{quote}

To bypass API restrictions, many researchers and small trust and safety vendors used open-source scrapers to collect historical user-generated content. However, these tools often misclassified romanized or code-mixed content as English due to the use of Latin scripts. Recently, companies like Meta, X, and Reddit have blocked scrapers and restricted API access to user content (e.g., Meta's shutdown of CrowdTangle)~\cite{Mehta-2023, Bellan-2024, Walker-2024, Perez-2024}. These restrictions, combined with limited datasets, have severely hindered efforts to study disinformation and hate speech in high-risk Global South regions. P18, a trust and safety practitioner at TikTok remarked:
\begin{quote}
    \textit{``After ChatGPT came out, companies are cautious of publicly sharing their data given the competition to develop their own language models. That's why we no longer see that openness around sharing data.''}
\end{quote}

Due to challenges in accessing and curating datasets, some researchers ceased studying online harms in their regions. Others started collecting online posts manually, using data from shared tasks at NLP conferences, deploying community surveys, or voluntary WhatsApp data donations. These methods were time-consuming and yielded small, inconsistent datasets, often insufficient for training AI models. 

Several participants emphasized the need for equitable access to user-generated content. While acknowledging privacy concerns, they criticized platforms like TikTok for limiting API access to researchers in the US and Europe~\cite{tiktok}. In response to such disparities, African researchers have joined grassroots initiatives like Masakhane and Tanzanian AI to reclaim ownership of locally generated data. Academics also condemned tech companies for mishandling harmful content in their regions while monopolizing user data. P19, who worked at Meta commented:
\begin{quote}
    \textit{``When I worked at Meta, the trust and safety team prioritized the US. These for-profit corporations derive most of their revenues from Western markets. Although Europe has strong regulatory policies, those markets are important to the company. So the prioritization simply reflects that.''}
\end{quote}

Participants raised concerns about how tech companies address data scarcity in low-resource languages. They criticized keyword-based filtering for overlooking dialectal variation and treating these languages as monolithic. They also flagged biases in machine-translated texts, often used as a workaround for limited data. For example, Kenyan Swahili researchers noted that Google Translate heavily supports outdated Sheng (a cerole combining Swahili and English) over its modern variants like Shembeteng, while Tanzanian researchers found it biased toward Kenyan Sheng over standard Tanzanian Swahili. Similarly, Quechua researchers highlighted tech company's problematic reliance on outdated Bible translations and colonial-era texts. One industry practitioner also noted their company using old Arabic dictionary due to scarce datasets in Maghrebi Arabic. Participants emphasized that models built on outdated corpora and biased machine translations are ill-suited to address the evolving nature of hate speech online. 

Small AI startups and trust and safety vendors reported that big tech companies often showed interest in their tools and datasets--only if offered for free. They demanded tech companies to support grassroots, local research efforts to address online harms in the Global South. P9, a Quechua linguist expressed: 
\begin{quote}
    \textit{``They [companies] should work with us, indigenous Quechua people, to build corpuses instead of taking the shortcut by using machine-translated texts. We found rule-based translation that incorporates grammatical knowledge works better for Quechua than stochastic methods, which require lots of data that do not exist in Quechua. When we contacted Google, they proposed us to work voluntarily. But I worry they will try to appropriate our free labor.''}
\end{quote}

These findings show that tech companies' lack of support for local data curation efforts and gatekeeping of user-generated content worsen data scarcity, disrupting grassroots efforts to tackle online harms in the Global South.

\subsection{Difficulties in Annotating Harmful Content}\label{annotation}
Annotation involves labeling the data to train AI models to detect harmful content and identify its type. Tech companies frequently outsource annotation to human moderators in the Global South. One industry practitioner noted that their company assigned Kenyan moderators to annotate Swahili dialects they didn’t understand. Their efforts to assign content to moderators with appropriate language expertise often fail due to poor language identification tools for low-resource languages. Participants stressed that companies have always underfunded annotation efforts for languages spoken in \textit{``less profitable regions.''} P19, who worked at a social media company, shared:

\begin{quote}
    \textit{``During Arab Spring, [redacted] had only two Arabic speaking moderators. There's so much diversity in the Arab world-- it's unlikely that the two moderators will get the full context of Arab Spring in Tunisia or Green Movement in Iran. Although a lot has changed since then, the core structure and issues remain the same.''}
\end{quote} 

Participants expressed that tech companies' lack of understanding of ground realities, socio-cultural norms, and linguistic nuances--significantly hinder their ability to address harmful content in the Global South. They stressed that tech companies should \textit{``give Global South a seat at the table''} when annotating hate speech. P12, a Kenyan researcher working on Swahili, remarked: 
\begin{quote}
    \textit{``It matters who is defining hate speech. We noticed that people use superlatives and `US' vs. `Them' narrative to spread supremacist views and ethnic hate speech. We developed our annotation framework to capture these cases. Since Twitter did not remove these tweets, their definition of hate speech must be different. By allowing these posts Twitter is reinforcing stereotypes about Africans being violent.''}
\end{quote}

To ensure that the annotated datasets capture local sensitivities, local researchers 
often involved linguists, activists, and affected communities to inform their annotation guidelines. P3, a researcher working on Tamil, explained:
\begin{quote}
    \textit{``It's important to consider intersectionality when annotating hate speech in multicultural environments like India, where caste, religion, and gender are intertwined. For example, we found `shuttlecock' [badminton cork] is used as a derogatory term against Muslim women who wear burka. Our team of feminist activists, experts on gender studies, and survivors of harassment helped us annotate coded hate speech that are both misogynist and Islamophobic. Similarly, there were innocuous comments like `you are my sweetheart.' When companies recruit gigworkers who are usually male, they would rate this as harmless. But since we worked with victims of sexual harassment and recipients of such comments, they could recognize these messages are part of broader harassment Indian women face online.''}
\end{quote}

Researchers valued involving community partners in data annotation but faced funding challenges that hindered sustained collaboration, annotator training, and quality control. Due to limited resources, they often relied on undergraduates to annotate hate speech and toxic content without being able to provide mental health support. P11, an academic researcher working on Quechua, shared:
\begin{quote}
    \textit{``Very often the dataset we are creating is the first of its kind in Quechua. Although experts and community members are willing to help voluntarily, it's difficult to sustain their free labor in the long run to annotate large volumes of data. So, we often strategize to annotate only a subset of data. We can't rush people to annotate faster because they are helping out of generosity. Thus, it takes months to annotate anything.''}
\end{quote}

To make the most of limited annotation resources, researchers often used sentiment and toxicity analysis tools to find negative content, reducing the sample size for manual annotation. However, they noted that existing free and proprietary tools from tech companies often lack cultural nuances. P13, a researcher specializing in Swahili, elaborated:

\begin{quote}
    \textit{``In America, people casually use the word `dawg' to refer to buddy but in Kenya calling someone dawg will be disrespectful. Similarly, in America people think calling `fat' is body shaming. In Africa fat is considered beautiful and opulent. But Google's perspective API missed these cases by applying American scale.''}
\end{quote}

Tamil researchers shared that they lose valuable annotation time and budget when manually verifying target languages in scraped corpuses because language identification tools have poor coverage for most low-resource languages. These tools often fail to separate code-mixed Tanglish (Tamil-English) from Kannada-English or Telugu-English because Tamil, Kannada, and Telugu often share words with same roots. Similarly, Maghrebi Arabic researchers reported that these AI tools often fail to differentiate between Arabic, Farsi, and Urdu due to overlapping scripts.

These findings reveal that limited AI support for low-resource, non-English languages hampers annotation efforts, which are further strained by chronic underfunding of grassroots initiatives. Despite having ample resources, tech companies frequently overlook cultural nuances of online harms due to insufficient engagement with stakeholders and communities in the Global South. 

\subsection{Preprocessing Challenges for Harmful Content Detection}\label{preprocessing}
Preprocessing involves cleaning and transforming raw data in a suitable format to train AI models. Our participants faced several challenges when applying existing preprocessing techniques on low-resource languages.

\parabold{Tokenization} 
Tokenization involves segmenting text into words or subwords for language processing. Participants noted that multilingual models like BERT and RoBERTa use frequency-based tokenization methods (e.g., WordPiece, BPE), generating tokens based on the frequency of words or co-occurring character pairs. These methods, primarily designed for English, underperform on languages like Tamil, Swahili, Maghrebi Arabic, and Quechua, which have more complex morphology than English. They explained that these languages have agglutinative properties, forming complex words by combining multiple morphemes (i.e., the smallest unit of meaning), where each morpheme retains its original meaning. For example, the Quechua word \textit{`rimanqakuma'} (meaning, they will definitely speak) consists of three morphemes: \textit{`rima-'} (meaning, to speak), \textit{`-nqa'} (refers to future tense) and \textit{`-kuma'} (signifies emphasis). The final meaning is directly derived from these constituent morphemes. 
P9 elaborated further stressing the need to derive morphemes correctly during tokenization: 
\begin{quote}
    \textit{``Frequency based tokenizers have been designed considering English as a model language. Since English is data-rich, frequency based method really works well. But for low-resource, agglutinative languages it creates illegible tokens by wrongly splitting the morphemes. If we train models with wrongly split tokens, the models won't derive correct embeddings. Instead, when we used linguistically motivated tokenizer, the performance significantly improved for Quechua in downstream tasks.''}
\end{quote}

Maghrebi Arabic NLP researchers also noted that using specialized morphological and monolingual tokenizers improve sentiment analyses for diverse low-resource languages, typically underrepresented in multilingual models. Swahili researchers further highlighted the challenges of tokenizing code-mixed hashtags that are often used to incite attacks while evading detection by platforms. For example, in the Sheng hashtag \textit{\#TupataneTuesday} (meaning, let's meet each other on Tuesday), used by protesters, the Swahili word \textit{Tupatane} must be correctly segmented into its morphemes: Tu- (we), -pat (to meet), -ane (each other). However, poor performance of language identification technologies on code-mixed texts complicates the application of tokenization algorithm based on language.

\parabold{Normalization}
Researchers also identified challenges in the normalization process, where words are converted to their standard forms before tokenizing (e.g., baaaad is normalized to bad). Some participants reported that non-standard spelling of agglutinative words causes confusion during normalization. P6, a Tamil researcher from Sri Lanka, explained:
\begin{quote}
    \textit{``In Tamil, `Amma' means Mother and `Ama' means Yes. On social media people often enthusiastically write Ama as `Aammaa' (similar to Yeessss) or distort the word Amma as `Aammaa' in gendered slurs. The model often makes errors while normalizing such cases and fails to flag offensive language.''}
\end{quote}

\parabold{Stemming and Lemmatization} These steps are performed to reduce words to their meaningful roots before training models (e.g., beautiful and beautify are reduced to beauty). Several participants reported facing challenges because existing tools have higher error rates in complex agglutinative languages, where \textit{``each root can take thousands of inflected forms''}, than morphologically simpler languages like English. P6 further described:

\begin{quote}
    \textit{``Both understemming and overstemming of complex Tamil grammar can cause error in detecting offensive language. Words like Mulaicchu (meaning, n**ples) often wrongly gets stemmed to Mulai- (meaning, sprout) and then gets ignored by model.''}
\end{quote}

\parabold{Parts-of-Speech (POS) Tagging} Some participants reported that since most models are trained on English, which is a subject-verb-object (SVO) language, it leads to errors on languages that follow subject-object-verb (SOV) structure. Therefore, they performed POS analysis during data preprocessing to give models additional contexts about derogatory adjectives and verbs aimed at individuals or groups (nouns). For example, in \textit{N\={a}yai serupp\={a}la a\d{t}ikka\d{n}um} (meaning, beat the dog with sandals) the object (noun) \textit{N\={a}yai} appears before the verb \textit{a\d{t}ikka\d{n}um}. However, researchers faced several challenges in detecting POS due to code-mixing. P2, an academic researcher, shared:

\begin{quote}
    \textit{``When I started doing NLP research in early 2000, there was no POS tagger for Tamil. There was barely any dataset to work with. We built corpuses from scratch and worked with linguists to annotate complex Tamil vocabulary. But the POS tagger based on monolingual Tamil does not work well on Tanglish from social media. Although many frame code-mixed data as problematic and low-quality, this is the reality of how social media users from non-English speaking countries write online. Handling code-mixing is very challenging. But we don't have access to code-mixed data from social media since they stopped access.''}
\end{quote}

These findings show that current preprocessing techniques, predominantly developed with English in mind, do not account for the morphologically rich and code-mixed nature of languages in the Global South, reflecting historical imbalances in linguistic and technological priorities.

\subsection{Challenges in Developing AI Models for Harmful Content Detection}\label{models}
After standard preprocessing, the data is fed into AI models for training and detecting harmful content. Participants reported using multilingual models, such as Google's mBERT, Facebook's XLM-RoBERTa, and AI4Bharat's IndicBERT but found these models perform poorly when detecting harmful content in low-resource languages. They explained that although these data-driven models are designed to be language-agnostic, being primarily trained on high-resource languages like English, they better learn the simpler morphology and fixed word orders of English. In contrast, data sparsity in low-resource languages limits these models' ability to capture the rich inflectional morphology, agglutinative property, complex grammar, and diverse word orders that are absent in English. P4, specializing in Tamil, described:
\begin{quote}
    \textit{``English and Tamil are from different language families and Tamil has richer morphology than English. How can these models derive correct embeddings of complex Tamil words by computing from the point of view of English? That's why IndicBERT doesn't perform well. There, Hindi and Marathi are from the same family but Tamil is a Dravidian language. So without considering the specifics of language families, you can't get performance improvement.''}
\end{quote}

Researchers cautioned that adding data from multiple languages can degrade model's performance in both low-resource and high-resource languages due to limited model capacity, a phenomenon known as the \textit{``curse of multilinguality''}~\cite{chang2023multilinguality}. Additionally, they criticized how AI models cannot handle diversity in annotations, especially for languages like Tamil and Swahili that have tens of dialectical variations. P15, a startup founder focusing on Swahili AI, explained:
\begin{quote}
    \textit{``In Swahili the word `right' has at least 20 different transliterations depending on the context. Similarly, in my region, the word `Mathikkalla' refers to `I could not recognize you' but in other regions, the same word means `to neglect someone.' So, annotators would label the same content differently depending on their region. This impacts offensive language detection because current approaches to train AI models flatten annotator diversity into a singular label.''}
\end{quote}

Some researchers observed that large language models frequently misclassify code-mixed content during hate speech detection, especially when the spelling and words signal non-Western ethnicity. Trust and safety practitioners attributed these errors to a lack of diversity within tech companies and shared that very often their teams are linguistically and culturally homogeneous. They commented that company's diversity efforts often end at recruitment; once hired, employees have to work following company’s priorities, which are typically centered around English. P20 remarked how this lack of diversity leads to biased models:
\begin{quote}
    \textit{``In Western media, Arabic phrases, such as `Allahu Akbar' [God is great] mostly appear in the context of terrorism. When companies train AI models on such articles, the models learn these negative associations. But there is none in these teams to inform that local people use these phrases to express everyday joy and sorrow, beyond the instances of extreme speech portrayed by Western media.''}
\end{quote}

Industry practitioners shared that despite the shortcomings of large language models in low-resource languages, their companies are prioritizing AI models over alternative linguistic approaches they used in the past. They emphasized the advantages of using AI models, particularly in reducing the burden of tedious and distressing moderation work for humans. In contrast, AI researchers and practitioners working in the Global South highlighted their struggle in training billion parameter models due to a lack of funding, computational power, and appropriate hardware. For example, Swahili researchers and engineers shared that they could not buy GPUs in Kenya and Tanzania and had to rely on their contacts in the US to access these resources. 
Many pointed out that free resources from Google Colab and Kaggle are barely enough to experiment with, train, and deploy these large language models. 
P12 commented:
\begin{quote}
    \textit{``We lack the necessary data, funding, and resources to build dedicated models for our languages. We spend time scraping for little data and cleaning it. I hope we can decolonize NLP research on online harms, so that we don't have to rely on technologies favoring high-resource languages like English and developed for nations with lots of computing power.''}
\end{quote}

These findings highlight that the resource-intensive and English-centric design of large language models is ill-equipped to address online harms in low-resource languages from the Global South, reflecting how the needs of these communities are usually sidelined in the development of AI-driven moderation technologies.

\section{Discussion}

While prior work attributes moderation challenges in low-resource languages to the lack of labeled datasets~\cite{Rowe-2022, nicholas-2023}, our study uncovers how socio-political factors in technology design exacerbate these issues. Our findings underscore how tech companies continue to rely on biased machine translation systems using outdated corpora instead of collaborating with experts and communities from the Global South—often appropriating their free labor. We reveal how these companies' blanket data restrictions for building proprietary large language models aggravate data scarcity to address online harms in these regions (\ref{dataset}). While prior studies report biased and opaque annotation practices among tech workers~\cite{Scheuerman-2024} and ML researchers~\cite{Geiger-2020}, we examine the structural factors enabling these issues. Socio-political issues, such as tech companies' weak financial incentives to improve annotation for Global South languages and limited funding available to Global South researchers along with technical issues like Western-centrism in sentiment and toxicity models and treating code-mixed data as ``poorer quality'' when developing language detection tools--compromise the annotation process (\ref{annotation}).

Moreover, most studies explain away preprocessing and model building challenges in low-resource languages by highlighting data scarcity~\cite{khan2023, zhong2024}. In contrast, our study questions the status quo that prioritizes data-intensive methods while overlooking alternative approaches that center linguistic diversity, morphological complexity, and dynamic evolution through code-mixing and code-switching—phenomena largely absent in English (\ref{preprocessing}, \ref{models}). We provide concrete examples of how normative assumptions in technology design contribute to moderation errors in diverse Global South languages--that remain invisible when assessed solely through low accuracy rates. Our focus on diverse languages help us establish the systemic nature of moderation biases. In discussion, we probe deeper into these systemic inequities, unpacking their historical and socio-political roots--often overlooked in existing discourse (\ref{sec:coloniality}). We then discuss approaches to improve moderation for low-resource languages while acknowledging the complexity of the issue (\ref{sec:paths}). 

\subsection{Coloniality in Moderation Pipelines}\label{sec:coloniality}

\parabold{Data Curation} Our findings show that tech companies lack interest to expend moderation resources for less profitable markets in the Global South. Our participants stressed that companies benefit by monopolizing user-generated data to train proprietary large language models, while restricting researchers' access to the very data needed for detecting harmful content. For instance, shortly after Reddit locked public data~\cite{Perez-2024}, it partnered with OpenAI to enable training ChatGPT on its content~\cite{openai-2024}. Similarly, Meta launched AI across Facebook, WhatsApp, and Instagram to train proprietary models on public posts without letting users opt out~\cite{Jesus-2024}, while simultaneously closing CrowdTangle that allowed researchers to access public content on Meta~\cite{Bellan-2024}. Researchers criticized these blanket restrictions on public data as privacy washing, impeding trust and safety scholarship within academia and civil society~\cite{Arney-2024}. 

These restrictions disproportionately affect researchers and practitioners in the Global South, where datasets in non-English languages remain scarce. This data scarcity stems from colonial legacy that suppressed Indigenous and native languages in the Global South~\cite{Thiong-1986, bank-2014, Otosirieze-2018, Kolli-2024} and deprioritized their digitization and technology development~\cite{bird-2020, schwartz-2022, held-2023, ogunremi-2023}. The systemic omission affects all downstream NLP tasks in low-resource languages, including automated moderation-- further hampered by data restriction imposed by tech companies.

Our participants highlighted that the data controlled by tech companies are generated through the unpaid labor of users in their communities. \citet{coleman-2018} explains that Facebook introduced Free Basics initiatives in the Global South to extract data from the region's next billion users, taking advantage of weak data protection laws and regulatory frameworks. \citet{kwet-2019} likens this process to digital colonialism. He argues that much like colonizers who built railroads to extract material resources from colonies, tech companies control digital infrastructures in the Global South, reduce local communities to products rather than producers, and commodify their data for corporate profit. 

Our analysis reveals that tech companies' reliance on cheap web-scraped data, machine translations, and religious texts for low-resource languages~\cite{kreutzer-2022, christodouloupoulos-2015, Ghosh-2023}--introduces significant biases in moderation. This includes wrongly associating Arabic phrases with terrorism and normalizing extreme speech in African contexts. Such biases reflect digital orientalism~\cite{alimardani-2021}, where colonial perspectives shape discriminatory narratives regarding the colonized `other'~\cite{said1977}. Likewise, the use of colonial-era texts to build Quechua datasets overlooks the historical role of colonial churches in suppressing Indigenous languages, while appropriating them only for cultural control~\cite[p.~29]{heller-2017}. Thus, our findings highlight how colonial legacies continue to shape the availability of data sources--required to study online harms in the Global South.

\parabold{Annotation} Trust and safety practitioners in our study noted that tech companies lack incentives to hire expert moderators for Global South content. Yet, these companies often outsource English-language annotation to the Global South, exploiting low wages and weak labor protections, and inflicting psychological harm to local moderators~\cite{Dwoskin-2019, Elswah-2024}. This practice mirrors colonial exploitation, where the Global South workforce serves the interests of the Global North with little regard for local needs or safety~\cite{Posada-2021, Malik-2022}. 

Additionally, we found that limited funding in Global South institutions hinders grassroots efforts to annotate harmful content in local languages. Historically, resources extracted through colonial exploitation enabled Western nations to advance their scientific agenda and build extensive datasets~\cite{schopf-2020}. Consequently, most misinformation research focuses on the West due to easy availability of annotated datasets in English~\cite{Rananga-2024}. These systemic inequities, marked by resource scarcity in the Global South and tech companies' disinterest in investing in these regions~\cite{nicholas-2023, deGregorio-2023}--further limit the availability of annotated datasets in low-resource languages.

\parabold{NLP Tools Used in Moderation} Our findings underscore that current NLP technologies, primarily designed for English, overlook the cultural context, linguistic complexity, and evolution of languages in the Global South. For example, our participants reported that Google's Perspective API misinterprets diverse notions of toxicity across different cultures. Similarly, \citet{Das-2024} demonstrate that sentiment analysis tools for low-resource languages disproportionately associate negative sentiment with certain religious and national identities-- replicating colonial hierarchies of division sowed by British rulers in the Indian subcontinent. 

Decolonial scholars and historians have long documented the colonial project of standardizing European languages by creating dictionaries and grammars to assimilate Indigenous populations while  suppressing local languages~\cite{fishman-1989, heller-2017, anderson-2020, fanon-2023}. These forced affected communities to code-switch between native and European languages to navigate colonized spaces~\cite{mufwene-2004}. These legacies resulted in poor early support for non-Latin scripts online, continuing to hinder participation from speakers of many low-resource languages~\cite{van-2019, held-2023, nigatu-2024}. This discrimination has forced non-English speakers to adopt romanization and code-mixing for communicating online~\cite{held-2023}. However, the closed, proprietary language models, relying on \textit{sanitized} datasets, disenfranchise local knowledge, impose Western normative values without empowering local communities to align the model to their own values, forestall alternative visions, and perpetuate colonial binaries  that frame advanced technologies as rescuing ``primitive'' languages~\cite{verran-2007, bird-2020, varshney2024decolonial}.

Primarily being trained on English, these language models perform well on languages that share important typological properties with English~\cite{bender-2009, arnett-2024}. Thus, these models fail to capture the elaborate morphology present in many low-resource languages. Historically, linguists considered agglutinative languages as \textit{``less evolved''} than Western languages, such as Spanish, Greek, or German~\cite{errington2007}. \citet{bender-2009} critiques AI models for making assumptions about language structures that advantage some languages at the expense of others, highlighting their inherent lack of language independence. Scholars criticize such one-size-fits-all solutions for embodying \textit{``colonial impulse''} that disregards the ecology of diverse languages and perpetuates colonial hierarchies~\cite{dourish-2012, bird2022}. For languages spoken in the Global South, this translates to collapsing their linguistic diversity and complexity to a simplistic construct of data scarcity-- often taken at the face value. 

In sum, our findings show that existing challenges affecting automatic detection of harmful content in low-resource languages are often systemic and run deeper than the mere availability of data. 

\subsection{Considerations for A Path Forward}\label{sec:paths}

Tackling harmful content in low-resource languages is a complex issue shaped by conflicting priorities across stakeholders. Tech companies often see investing in moderation systems for these languages as \textit{unprofitable} despite these languages having millions of speakers~\cite{nicholas-2023, deGregorio-2023}. The deprioritization of trust and safety efforts within US-based tech companies undermines global accountability, reinforcing a US-centric vision of free speech~\cite{Scarcella-2024, Divon-2025}. Researchers also face little incentive to build labeled datasets for low-resource languages due to high effort~\cite{Sambasivan-2021} and limited academic payoffs~\cite{held-2023}. Governments in many Global South countries often resort to censorship or data localization laws due to platforms' failure to address harmful content but end up facing pushback from US tech lobbies~\cite{Kak-2020, deGregorio-2023}. Meanwhile, civil society groups in the Global South frequently feel marginalized as tech companies often treat local collaboration as a checkbox exercise rather than genuine partnership~\cite{bhrrc-2024}. Acknowledging these realities, we outline some concrete steps to make content moderation more equitable.

\parabold{Strengthening Local Research Capacity} Prior research highlights that when Global North institutions are funded to develop models for low-resource languages without involving local experts, they often fail in context-specific moderation tasks~\cite{nicholas-2023}. \citet{bhabha2011our} argues that enhancing national resources of the Global South is essential to addressing the geo-politics of resource distribution and facilitate redistributive justice. Therefore, governments, grant-making agencies, and research award programs by tech companies must invest in building self-sustaining, grassroots research ecosystems that actively engage local experts from the Global South. For example, the AI4D Africa program, funded by international governments and research institutes, supports the development of local AI research hubs and talent, empowering African researchers to lead projects that address their communities' needs~\cite{IDRC-2024}. Initiatives like Masakhane in Africa, AI4Bharat in India, and ARBML in the Arab World, which are democratizing AI research on low-resource languages, should be strengthened through targeted funding to amplify their impact. 

\parabold{Labeled Datasets} Although social media companies frequently cite privacy concerns in data sharing, established practices from other fields suggest feasible solutions. For instance, the Yale Open Data Access Project allows medical companies to securely share anonymized clinical trial data with vetted researchers for approved studies~\cite{nicholas-2022}. Similarly, differential privacy techniques can protect personal information when sharing large datasets~\cite{Kapelke-2020, Garfinkel-2022}. Since for-profit companies have little incentive to share user data, policy measures like the Digital Services Act (DSA Articles 12 and 40), which grant vetted researchers access to data from very large online platforms~\cite{Commission-2025}, can offer a potential solution. The DSA should seriously consider the data needs of researchers building automated tools to address online harms in low-resource languages. 

Additionally, for languages with a significant digital presence, voluntary data donation by native speakers can be useful for grassroots researchers. For example, \citet{garimella-2024} developed a data donation tool for closed WhatsApp groups while safeguarding the privacy of both donors and their contacts. In contrast, for Indigenous languages with limited digital presence, building respectful and equitable community relationships is essential, prioritizing local agency in community partnerships~\cite{bird-2020, Thakur-2025}. Indigenous Protocol and AI working group has created guidelines on how to integrate Indigenous ontology and epistemology when building AI tools~\cite{Lewis-2020} that can educate AI practitioners to not rely on colonial epistemology (e.g., Bible data) for Indigenous languages. 

Participants emphasized the need for diverse perspectives when annotating intersectional hate speech, urging companies to prioritize high recall to better capture harmful content~\cite{parrish-etal-2024-diversity}. For instance, during the Tigray conflict, moderators from dominant ethnic groups influenced how ethnic hate speech got labeled, while minority moderators--often at the bottom of the hierarchy--had limited influence on platform policy~\cite{Iyer-2025}. Since for-profit motives often discourage companies from diversifying moderators, governments can apply regulatory pressure--similar to data localization laws--to require platforms to recruit local moderators for content from their respective regions. For example, India's Information Technology (IT) Rules, 2021 mandate major social media companies to appoint local grievance officers and compliance staff based in India, effectively pressuring companies to localize their moderation teams and workflows~\cite{it-rule-2021}.

Moreover, it is important to follow the best practices for supporting community labor when annotating harmful content by disclosing the task, offering opt-out options, providing well-being support, and monetary compensation ~\cite{Evani-2024}. For example, Karya---a nonprofit data company based in India---empowers disadvantaged communities through data annotation work and pays them nearly 20 times more than the local minimum wage~\cite{Perrigo-2023}. Local labor rights organizations can partner with initiatives like the Global Trade Union Alliance of Content Moderators~\cite{union-2025} to advocate for fair treatment of moderators. Together, they can pressure tech companies to balance workloads when assigning traumatizing content and to provide fair wages, mental health support, and safer working conditions.~\cite{Elswah-2024}.

\parabold{Language-Aware Solutions} Since current NLP tools and language models are inherently English-centric, our participants recommended approaches that incorporate linguistic knowledge, such as using morphological segmenters instead of frequency-based tokenizers~\cite{abdelali2021pre, zevallos-bel-2023-hints}, rule-based translations over stochastic machine translations~\cite{sreelekha-2018}, and vector embeddings of local hateful phrases for detecting code-mixed hate speech~\cite{Devi-2024}. Some participants also discouraged using multilingual models for moderation given these models fail to infer correct linguistic context for different language families. However, given the arms race among tech companies to develop large multilingual models~\cite{Gupta-2024}, it is unlikely that they will shift to such linguistically informed solutions without regulatory pressures. Meanwhile, limited access to computing power in the Global South limits researchers' capacity in training and experimenting with \textit{``language-specific''} (monolingual) models and \textit{``language-aware''} approaches that do not necessarily rely on vast datasets or huge computing power. Free computing resources provided by tech companies, such as Google's Colab and TPU Research Cloud programs--remain inadequate for research purposes. Expanding access to these resources is critical to enable more equitable and inclusive moderation research in Global South languages. 

\parabold{Policy and Practice} For the data and tools created by local experts to have meaningful impact, they must be deployed to moderate local content. Cohere's partnership with HausaNLP to integrate African language datasets into its multilingual Aya model~\cite{Evani-2024} demonstrates the potential of such efforts. Similarly, the scales created by local researchers to evaluate model's performance for detecting code-mixed hate speech~\cite{das2022}--should be integrated into tech companies' evaluation frameworks. Governments and civil society should create regulatory policies that require tech companies to prioritize local representation, data ownership, and community self-determination. These policies must articulate the specific harms caused by flawed content moderation systems in high-risk, under-resourced contexts, rather than simply applying Western frameworks of harm to regional trust and safety efforts~\cite{Kennedy-2025}. Additionally, they should push for the inclusion of model performance metrics for low-resource languages in transparency reports, evaluated against locally defined benchmarks. Given the limitations of accuracy metrics in class-imbalanced content moderation task, regulatory frameworks should mandate the inclusion of more informative measures such as recall (\% of correctly flagged harmful content) and precision (\% of flagged content that is truly harmful)~\cite{Wei-2025}. Such steps could surface the limitations of current models and incentivize progress toward better moderation of harmful content in underrepresented languages.



\section{Conclusion}

Content moderation is difficult but even more so for low-resource languages due to systemic issues and colonial legacies impacting data access, culturally-aware annotation, and the incentive to build language-aware technologies. As a result, the techno-solutionist approaches championed by Silicon Valley fail to address the multifaceted complexities of detecting online harms in low-resource languages. To break this vicious cycle, we need a bottom-up approach that invests in digital public infrastructure and prioritizes online safety over corporate profit.

\section{Positionality Statement}

All authors of this work come from historically colonized regions in the Global South and are native speakers of languages, which are considered \textit{``low-resourced.''} All authors have extensive experience of doing critical research in diverse Global South contexts. Although none of the authors are affiliated with industry, our background in academia (computer science and communications) and civil society---enabled us to engage with participants both at technical and policy levels. Although we come from historically colonized countries, we are affiliated with academic institutions that benefited from colonial expansion and were established using wealth derived from the forced appropriation of Indigenous lands. We acknowledge that these affiliations afford us research privileges, such as access to funding, institutional support, and global visibility that are often inaccessible to many Global South researchers. In this tension, we identify with \citet{villenas1996colonizer}'s notion of having \textit{``feet in both worlds,''} as we simultaneously belong to communities shaped by colonial histories and to academic institutions that have profited from those same colonial legacies.

Similarly, we recognize that caste, religion, ethnicity, and other intersectional identity likely shaped the experiences of our participants although we did not explicitly collect such data. For instance, several Indian participants studying Tamil were likely from upper-caste backgrounds, some researchers studying Quechua were not of Indigenous origin, and many had greater access to resources due to Western affiliations compared to their counterparts in the Global South.

Following \citet{fine-1994working}'s self-reflexive approach, we acknowledge the intersections of privilege and marginalization among our participants. For instance, Quechua-speaking Indigenous researchers reported greater struggle in practicing their native language in academic spaces compared to Western researchers studying the same Indigenous language. Similarly, the experiences of upper-caste AI experts would differ from those coming from lower-caste backgrounds–who might face additional challenges and lack social capital in pursuing AI research in Tamil. Swahili researchers and practitioners from Tanzania felt their Kenyan counterparts enjoyed more visibility since most tech companies have regional offices within Kenya. Although industry practitioners had greater access to computing resources, they reflected on their limitations within corporate financial infrastructures and lack of diversity within these organizations. As \citet{haraway-2013} argues, knowledge production is always situated; the experiences of AI researchers and practitioners in our study form a \textit{``partial perspective''}–shaped by their backgrounds and the struggle of researching marginalized, low-resource languages.

\section{Acknowledgment}
We are earnestly thankful to Internet Society (ISOC), Dhanaraj Thakur, Aliya Bhatia, Cornell Global AI Initiative, Infosys, DeVan Hankerson Madrigal, Gabrierl Nicholas, and Sarah Zolad for supporting this work.

\bibliography{00_references}

\begin{thebibliography}{115}
\providecommand{\natexlab}[1]{#1}

\bibitem[{Abdelali et~al.(2021)Abdelali, Hassan, Mubarak, Darwish, and Samih}]{abdelali2021pre}
Abdelali, A.; Hassan, S.; Mubarak, H.; Darwish, K.; and Samih, Y. 2021.
\newblock Pre-training bert on arabic tweets: Practical considerations.

\bibitem[{Ahmad and Krzywdzinski(2022)}]{ahmad-2022}
Ahmad, S.; and Krzywdzinski, M. 2022.
\newblock \emph{Moderating in obscurity: How Indian content moderators work in global content moderation value chains}, chapter~5, 77--95.
\newblock Cambridge, MA: The MIT Press.

\bibitem[{Ali(2016)}]{ali2016brief}
Ali, S.~M. 2016.
\newblock A brief introduction to decolonial computing.
\newblock \emph{XRDS: Crossroads, The ACM Magazine for Students}, 22(4): 16--21.

\bibitem[{Alimardani and Elswah(2021)}]{alimardani-2021}
Alimardani, M.; and Elswah, M. 2021.
\newblock Digital orientalism:\# SaveSheikhJarrah and Arabic content moderation.

\bibitem[{Anderson(2020)}]{anderson-2020}
Anderson, B. 2020.
\newblock Imagined communities: Reflections on the origin and spread of nationalism.
\newblock In \emph{The new social theory reader}, 282--288. United Kingdom: Routledge.

\bibitem[{Arnett and Bergen(2024)}]{arnett-2024}
Arnett, C.; and Bergen, B.~K. 2024.
\newblock Why do language models perform worse for morphologically complex languages?

\bibitem[{Arney(2024)}]{Arney-2024}
Arney, J. 2024.
\newblock Data dump: Meta killed CrowdTangle. What does it mean for researchers, reporters?

\bibitem[{Bank(2014)}]{bank-2014}
Bank, W. 2014.
\newblock Discriminated against for speaking their own language.

\bibitem[{Bellan(2024)}]{Bellan-2024}
Bellan, R. 2024.
\newblock Meta axed CrowdTangle, a tool for tracking disinformation. Critics claim its replacement has just `1\% of the features'.

\bibitem[{Bender(2009)}]{bender-2009}
Bender, E.~M. 2009.
\newblock Linguistically Na{\"i}ve != Language Independent: Why {NLP} Needs Linguistic Typology.
\newblock In \emph{Proceedings of the {EACL} 2009 Workshop on the Interaction between Linguistics and Computational Linguistics: Virtuous, Vicious or Vacuous?}, 26--32. Athens, Greece: Association for Computational Linguistics.

\bibitem[{Benjamin(2023)}]{benjamin2023race}
Benjamin, R. 2023.
\newblock Race after technology.
\newblock In \emph{Social Theory Re-Wired}, 405--415. New York: Routledge.

\bibitem[{Bhabha(2011)}]{bhabha2011our}
Bhabha, H.~K. 2011.
\newblock \emph{Our neighbours, ourselves: Contemporary reflections on survival}.
\newblock De Gruyter.

\bibitem[{Bhattacharyya(2018)}]{bhattacharyya-2018}
Bhattacharyya, G. 2018.
\newblock \emph{Rethinking racial capitalism: Questions of reproduction and survival}.
\newblock Maryland, USA: Rowman \& Littlefield.

\bibitem[{Biddle(2022)}]{Biddle-2022}
Biddle, S. 2022.
\newblock Facebook's Tamil Censorship Highlights Risks to Everyone.

\bibitem[{Bird(2020)}]{bird-2020}
Bird, S. 2020.
\newblock Decolonising speech and language technology.
\newblock In \emph{28th International Conference on Computational Linguistics, COLING 2020}, 3504--3519. online: Association for Computational Linguistics (ACL).

\bibitem[{Bird(2022)}]{bird2022}
Bird, S. 2022.
\newblock Local languages, third spaces, and other high-resource scenarios.
\newblock In \emph{60th Annual Meeting of the Association for Computational Linguistics, ACL 2022}, 7817--7829. Dublin: Association for Computational Linguistics (ACL).

\bibitem[{Braun and Clarke(2006)}]{braun-2006}
Braun, V.; and Clarke, V. 2006.
\newblock Using thematic analysis in psychology.
\newblock \emph{Qualitative research in psychology}, 3(2): 77.

\bibitem[{Centre(2024)}]{bhrrc-2024}
Centre, B. . H. R.~R. 2024.
\newblock Dismantling the facade: A global south perspective on the state of engagement with tech companies.

\bibitem[{Chang et~al.(2023)Chang, Arnett, Tu, and Bergen}]{chang2023multilinguality}
Chang, T.~A.; Arnett, C.; Tu, Z.; and Bergen, B.~K. 2023.
\newblock When is multilinguality a curse? language modeling for 250 high-and low-resource languages.

\bibitem[{Christodouloupoulos and Steedman(2015)}]{christodouloupoulos-2015}
Christodouloupoulos, C.; and Steedman, M. 2015.
\newblock A massively parallel corpus: the bible in 100 languages.
\newblock \emph{Language resources and evaluation}, 49: 375--395.

\bibitem[{Coleman(2018)}]{coleman-2018}
Coleman, D. 2018.
\newblock Digital colonialism: The 21st century scramble for Africa through the extraction and control of user data and the limitations of data protection laws.
\newblock \emph{Michigan Journal of Race and Law}, 24: 417--439.

\bibitem[{Commission(2025)}]{Commission-2025}
Commission, E. 2025.
\newblock Commission adopts delegated act on data access under the Digital Services Act.

\bibitem[{Couldry and Mejias(2019)}]{couldry-2019}
Couldry, N.; and Mejias, U.~A. 2019.
\newblock Data colonialism: Rethinking big data's relation to the contemporary subject.
\newblock \emph{Television \& New Media}, 20(4): 336--349.

\bibitem[{Das et~al.(2024)Das, Guha, Brubaker, and Semaan}]{Das-2024}
Das, D.; Guha, S.; Brubaker, J.~R.; and Semaan, B. 2024.
\newblock The ``Colonial Impulse" of Natural Language Processing: An Audit of Bengali Sentiment Analysis Tools and Their Identity-based Biases.
\newblock In \emph{Proceedings of the 2024 CHI Conference on Human Factors in Computing Systems}. New York, USA: ACM.

\bibitem[{Das et~al.(2022)Das, Saha, Mathew, and Mukherjee}]{das2022}
Das, M.; Saha, P.; Mathew, B.; and Mukherjee, A. 2022.
\newblock Hatecheckhin: Evaluating hindi hate speech detection models.

\bibitem[{De~Gregorio and Stremlau(2023)}]{deGregorio-2023}
De~Gregorio, G.; and Stremlau, N. 2023.
\newblock Inequalities and content moderation.
\newblock \emph{Global Policy}, 14(5): 870--879.

\bibitem[{Devi, Kannimuthu, and Madasamy(2024)}]{Devi-2024}
Devi, V.~S.; Kannimuthu, S.; and Madasamy, A.~K. 2024.
\newblock The Effect of Phrase Vector Embedding in Explainable Hierarchical Attention-Based Tamil Code-Mixed Hate Speech and Intent Detection.
\newblock \emph{IEEE Access}, 12(0): 11316--11329.

\bibitem[{Divon and Ong(2025)}]{Divon-2025}
Divon, T.; and Ong, J.~C. 2025.
\newblock Tech Bro Power Play: Zuckerberg vs. Global Tech Justice.

\bibitem[{Dourish and Mainwaring(2012)}]{dourish-2012}
Dourish, P.; and Mainwaring, S.~D. 2012.
\newblock Ubicomp's colonial impulse.
\newblock In \emph{Proceedings of the 2012 ACM conference on ubiquitous computing}, 133--142. New York, USA: ACM.

\bibitem[{Elizabeth~Dwoskin and Cabato(2019)}]{Dwoskin-2019}
Elizabeth~Dwoskin, J.~W.; and Cabato, R. 2019.
\newblock Content moderators at YouTube, Facebook and Twitter see the worst of the web — and suffer silently.

\bibitem[{Elswah(2024{\natexlab{a}})}]{Elswah-2024}
Elswah, M. 2024{\natexlab{a}}.
\newblock Moderating Kiswahili Content on Social Media.

\bibitem[{Elswah(2024{\natexlab{b}})}]{Elswah-arabic-2024}
Elswah, M. 2024{\natexlab{b}}.
\newblock Moderating Maghrebi Arabic Content on Social Media.

\bibitem[{Errington(2007)}]{errington2007}
Errington, J. 2007.
\newblock \emph{Linguistics in a colonial world: A story of language, meaning, and power}.
\newblock John Wiley \& Sons.

\bibitem[{Fanon(2023)}]{fanon-2023}
Fanon, F. 2023.
\newblock Black skin, white masks.
\newblock In \emph{Social theory re-wired}, 355--361. United Kingdom: Routledge.

\bibitem[{Fine(1994)}]{fine-1994working}
Fine, M. 1994.
\newblock Working the hyphens.
\newblock \emph{Handbook of qualitative research}, 2.

\bibitem[{Fishman(1989)}]{fishman-1989}
Fishman, J.~A. 1989.
\newblock \emph{Language and ethnicity in minority sociolinguistic perspective}.
\newblock United Kingdom: Multilingual Matters.

\bibitem[{Garfinkel and Bowen(2022)}]{Garfinkel-2022}
Garfinkel, S.~L.; and Bowen, C.~M. 2022.
\newblock Preserving Privacy While Sharing Data.

\bibitem[{Garimella and Chauchard(2024)}]{garimella-2024}
Garimella, K.; and Chauchard, S. 2024.
\newblock WhatsApp Explorer: A Data Donation Tool To Facilitate Research on WhatsApp.

\bibitem[{Geiger et~al.(2020)Geiger, Yu, Yang, Dai, Qiu, Tang, and Huang}]{Geiger-2020}
Geiger, R.~S.; Yu, K.; Yang, Y.; Dai, M.; Qiu, J.; Tang, R.; and Huang, J. 2020.
\newblock Garbage in, garbage out? do machine learning application papers in social computing report where human-labeled training data comes from?
\newblock In \emph{Proceedings of the 2020 Conference on Fairness, Accountability, and Transparency}, FAT* '20, 325–336. New York, NY, USA: ACM.

\bibitem[{Ghosh and Caliskan(2023)}]{Ghosh-2023}
Ghosh, S.; and Caliskan, A. 2023.
\newblock ChatGPT Perpetuates Gender Bias in Machine Translation and Ignores Non-Gendered Pronouns: Findings across Bengali and Five other Low-Resource Languages.
\newblock In \emph{Proceedings of the 2023 AAAI/ACM Conference on AI, Ethics, and Society}, AIES '23, 901–912. New York, NY, USA: ACM.

\bibitem[{Gorwa(2019)}]{gorwa-2019}
Gorwa, R. 2019.
\newblock What is platform governance?
\newblock \emph{Information, communication \& society}, 22(6): 854--871.

\bibitem[{Gramsci(2020)}]{gramsci-2020}
Gramsci, A. 2020.
\newblock Selections from the prison notebooks.
\newblock In \emph{The applied theatre reader}, 141--142. New York, USA: Routledge.

\bibitem[{Gupta(2024)}]{Gupta-2024}
Gupta, S. 2024.
\newblock The AI arms race: Which LLMs are winning the enterprise battlefield?

\bibitem[{Haraway(2013)}]{haraway-2013}
Haraway, D. 2013.
\newblock Situated knowledges: The science question in feminism and the privilege of partial perspective 1.
\newblock \emph{Women, science, and technology}, 1: 455--472.

\bibitem[{Held et~al.(2023)Held, Harris, Best, and Yang}]{held-2023}
Held, W.; Harris, C.; Best, M.; and Yang, D. 2023.
\newblock A material lens on coloniality in nlp.

\bibitem[{Heller and McElhinny(2017)}]{heller-2017}
Heller, M.; and McElhinny, B. 2017.
\newblock \emph{Language, capitalism, colonialism: Toward a critical history}.
\newblock Canada: University of Toronto Press.

\bibitem[{IDRC(2024)}]{IDRC-2024}
IDRC. 2024.
\newblock Artificial Intelligence for Development.

\bibitem[{Irani et~al.(2010)Irani, Vertesi, Dourish, Philip, and Grinter}]{Irani-2010}
Irani, L.; Vertesi, J.; Dourish, P.; Philip, K.; and Grinter, R.~E. 2010.
\newblock Postcolonial computing: a lens on design and development.
\newblock In \emph{Proceedings of the SIGCHI Conference on Human Factors in Computing Systems}, CHI '10, 1311–1320. New York, NY, USA: ACM.

\bibitem[{Iyer(2025)}]{Iyer-2025}
Iyer, P. 2025.
\newblock What a New Study Reveals About Content Moderation in Tigray.

\bibitem[{Jim\'{e}nez(2024)}]{Jesus-2024}
Jim\'{e}nez, J. 2024.
\newblock Worried About Meta Using Your Instagram to Train Its A.I.? Here's What to Know.

\bibitem[{Kak(2020)}]{Kak-2020}
Kak, A. 2020.
\newblock ``The Global South is everywhere, but also always somewhere'': National Policy Narratives and AI Justice.
\newblock In \emph{Proceedings of the AAAI/ACM Conference on AI, Ethics, and Society}, AIES '20, 307–312. New York, NY, USA: ACM.

\bibitem[{Kapelke(2020)}]{Kapelke-2020}
Kapelke, C. 2020.
\newblock Using differential privacy to harness big data and preserve privacy.

\bibitem[{Kennedy and Campos(2025)}]{Kennedy-2025}
Kennedy, W.~M.; and Campos, D.~V. 2025.
\newblock Vernacularizing Taxonomies of Harm is Essential for Operationalizing Holistic AI Safety.
\newblock In \emph{Proceedings of the 2024 AAAI/ACM Conference on AI, Ethics, and Society}, AIES '25, 698–710. New York, NY, USA: ACM.

\bibitem[{Khan et~al.(2023)Khan, Ullah, Alharbi, Alferaidi, Alharbi, Yadav, Alsharabi, and Ahmad}]{khan2023}
Khan, M.; Ullah, K.; Alharbi, Y.; Alferaidi, A.; Alharbi, T.~S.; Yadav, K.; Alsharabi, N.; and Ahmad, A. 2023.
\newblock Understanding the research challenges in low-resource language and linking bilingual news articles in multilingual news archive.
\newblock \emph{Applied Sciences}, 13(15): 8566.

\bibitem[{Kolli(2024)}]{Kolli-2024}
Kolli, V. 2024.
\newblock Linguistic Colonialism: Moroccan Education and its Dark Past.

\bibitem[{Kreutzer et~al.(2022)Kreutzer, Caswell, Wang, Wahab, van Esch, Ulzii-Orshikh, Tapo, Subramani, Sokolov, Sikasote et~al.}]{kreutzer-2022}
Kreutzer, J.; Caswell, I.; Wang, L.; Wahab, A.; van Esch, D.; Ulzii-Orshikh, N.; Tapo, A.; Subramani, N.; Sokolov, A.; Sikasote, C.; et~al. 2022.
\newblock Quality at a glance: An audit of web-crawled multilingual datasets.
\newblock \emph{Transactions of the Association for Computational Linguistics}, 10: 50--72.

\bibitem[{Kwet(2019)}]{kwet-2019}
Kwet, M. 2019.
\newblock Digital colonialism: US empire and the new imperialism in the Global South.
\newblock \emph{Race \& Class}, 60(4): 3--26.

\bibitem[{Legon and Alsalman(2020)}]{Legon-2020}
Legon, A.; and Alsalman, A. 2020.
\newblock How Facebook can Flatten the Curve of the Coronavirus Infodemic.
\newblock Technical report, Avaaz.

\bibitem[{Lewis et~al.(2020)Lewis, Abdilla, Arista, Baker, Benesiinaabandan, Brown, Cheung, Coleman, Cordes, Davison, Duncan, Garzon, Harrell, Jones, Kealiikanakaoleohaililani, Kelleher, Kite, Lagon, Leigh, Levesque, Mahelona, Moses, Nahuewai, Noe, Olson, Parker~Jones, Running~Wolf, Running~Wolf, Silva, Fragnito, and Whaanga}]{Lewis-2020}
Lewis, J.~E.; Abdilla, A.; Arista, N.; Baker, K.; Benesiinaabandan, S.; Brown, M.; Cheung, M.; Coleman, M.; Cordes, A.; Davison, J.; Duncan, K.; Garzon, S.; Harrell, D.~F.; Jones, P.-L.; Kealiikanakaoleohaililani, K.; Kelleher, M.; Kite, S.; Lagon, O.; Leigh, J.; Levesque, M.; Mahelona, K.; Moses, C.; Nahuewai, I.~I.; Noe, K.; Olson, D.; Parker~Jones, {\=O}.; Running~Wolf, C.; Running~Wolf, M.; Silva, M.; Fragnito, S.; and Whaanga, H. 2020.
\newblock Indigenous Protocol and Artificial Intelligence Position Paper.
\newblock Project Report 10.11573/spectrum.library.concordia.ca.00986506, Aboriginal Territories in Cyberspace, Honolulu, HI.

\bibitem[{Malik(2022)}]{Malik-2022}
Malik, S. 2022.
\newblock Global labor chains of the western AI.

\bibitem[{Mehta(2023)}]{Mehta-2023}
Mehta, I. 2023.
\newblock X updates its terms to ban crawling and scraping.

\bibitem[{MeitY(2023)}]{it-rule-2021}
MeitY. 2023.
\newblock The Information Technology (Intermediary Guidelines and Digital Media Ethics Code) Rules, 2021.

\bibitem[{Milmo(2021)}]{Milmo-2021}
Milmo, D. 2021.
\newblock Rohingya sue Facebook for \pounds150bn over Myanmar genocide.

\bibitem[{Mohamed, Png, and Isaac(2020)}]{mohamed-2020}
Mohamed, S.; Png, M.-T.; and Isaac, W. 2020.
\newblock Decolonial AI: Decolonial theory as sociotechnical foresight in artificial intelligence.
\newblock \emph{Philosophy \& Technology}, 33: 659--684.

\bibitem[{Mufwene(2004)}]{mufwene-2004}
Mufwene, S.~S. 2004.
\newblock \emph{The ecology of language evolution}.
\newblock United Kingdom: Cambridge University Press.

\bibitem[{Nicholas and Bhatia(2023)}]{nicholas-2023}
Nicholas, G.; and Bhatia, A. 2023.
\newblock Toward Better Automated Content Moderation in Low-Resource Languages.
\newblock \emph{Journal of Online Trust and Safety}, 2(1).

\bibitem[{Nicholas and Thakur(2022)}]{nicholas-2022}
Nicholas, G.; and Thakur, D. 2022.
\newblock Learning to Share: Lessons on Data-Sharing from Beyond Social Media.

\bibitem[{Nigatu and Raji(2024)}]{Nigatu-2024-YT}
Nigatu, H.~H.; and Raji, I.~D. 2024.
\newblock ``I Searched for a Religious Song in Amharic and Got Sexual Content Instead'': Investigating Online Harm in Low-Resourced Languages on YouTube.
\newblock In \emph{Proceedings of the 2024 ACM Conference on Fairness, Accountability, and Transparency}, FAccT '24, 141–160. New York, NY, USA: ACM.

\bibitem[{Nigatu et~al.(2024)Nigatu, Tonja, Rosman, Solorio, and Choudhury}]{nigatu-2024}
Nigatu, H.~H.; Tonja, A.~L.; Rosman, B.; Solorio, T.; and Choudhury, M. 2024.
\newblock The Zeno's Paradox ofLow-Resource'Languages.

\bibitem[{Obi-Young(2018)}]{Otosirieze-2018}
Obi-Young, O. 2018.
\newblock Bantu's Swahili, or How to Steal a Language from Africa | Kamau Muiga.

\bibitem[{{\`O}g{\'u}nr\.{e}\`{m}{\'\i}, Nekoto, and Samuel(2023)}]{ogunremi-2023}
{\`O}g{\'u}nr\.{e}\`{m}{\'\i}, T.; Nekoto, W.~O.; and Samuel, S. 2023.
\newblock Decolonizing nlp for "low-resource languages": Applying abebe birhane's relational ethics.

\bibitem[{OpenAI(2024)}]{openai-2024}
OpenAI. 2024.
\newblock OpenAI and Reddit Partnership.

\bibitem[{Ovalle et~al.(2023)Ovalle, Subramonian, Gautam, Gee, and Chang}]{Ovalle-2023}
Ovalle, A.; Subramonian, A.; Gautam, V.; Gee, G.; and Chang, K.-W. 2023.
\newblock Factoring the Matrix of Domination: A Critical Review and Reimagination of Intersectionality in AI Fairness.
\newblock In \emph{Proceedings of the 2023 AAAI/ACM Conference on AI, Ethics, and Society}, AIES '23, 496–511. New York, NY, USA: ACM.

\bibitem[{Parrish et~al.(2024)Parrish, Prabhakaran, Aroyo, D{\'i}az, Homan, Serapio-Garc{\'i}a, Taylor, and Wang}]{parrish-etal-2024-diversity}
Parrish, A.; Prabhakaran, V.; Aroyo, L.; D{\'i}az, M.; Homan, C.~M.; Serapio-Garc{\'i}a, G.; Taylor, A.~S.; and Wang, D. 2024.
\newblock Diversity-Aware Annotation for Conversational {AI} Safety.
\newblock In Dinkar, T.; Attanasio, G.; Cercas~Curry, A.; Konstas, I.; Hovy, D.; and Rieser, V., eds., \emph{Proceedings of Safety4ConvAI: The Third Workshop on Safety for Conversational AI @ LREC-COLING 2024}, 8--15. Torino, Italia: ELRA and ICCL.

\bibitem[{Perez(2024)}]{Perez-2024}
Perez, S. 2024.
\newblock Reddit locks down its public data in new content policy, says use now requires a contract.

\bibitem[{Perrigo(2023)}]{Perrigo-2023}
Perrigo, B. 2023.
\newblock The Workers Behind AI Rarely See Its Rewards. This Indian Startup Wants to Fix That.

\bibitem[{Popli(2021)}]{Popli-2021}
Popli, N. 2021.
\newblock The 5 Most Important Revelations From the `Facebook Papers'.

\bibitem[{Posada(2021)}]{Posada-2021}
Posada, J. 2021.
\newblock The Coloniality of Data Work in Latin America.
\newblock In \emph{Proceedings of the 2021 AAAI/ACM Conference on AI, Ethics, and Society}, AIES '21, 277–278. New York, NY, USA: ACM.

\bibitem[{Quijano(2000)}]{quijano-2000}
Quijano, A. 2000.
\newblock Coloniality of power and Eurocentrism in Latin America.
\newblock \emph{International sociology}, 15(2): 215--232.

\bibitem[{Quijano(2007{\natexlab{a}})}]{quijano-2007}
Quijano, A. 2007{\natexlab{a}}.
\newblock Coloniality and modernity/rationality.
\newblock \emph{Cultural studies}, 21(2-3): 168--178.

\bibitem[{Quijano(2007{\natexlab{b}})}]{quijano-2007questioning}
Quijano, A. 2007{\natexlab{b}}.
\newblock Questioning ``race''.
\newblock \emph{Socialism and democracy}, 21(1): 45--53.

\bibitem[{Radiya-Dixit and Bogen(2024)}]{Evani-2024}
Radiya-Dixit, E.; and Bogen, M. 2024.
\newblock Beyond English-Centric AI Lessons on Community Participation from Non-English NLP Groups.

\bibitem[{Rananga et~al.(2024)Rananga, Isong, Modupe, and Marivate}]{Rananga-2024}
Rananga, S.; Isong, B.; Modupe, A.; and Marivate, V. 2024.
\newblock Misinformation Detection: A Review for High and Low-Resource Languages.
\newblock \emph{Journal of Information Systems and Informatics}, 6(4): 2892--2922.

\bibitem[{Rowe(2022)}]{Rowe-2022}
Rowe, J. 2022.
\newblock Marginalised languages and the content moderation challenge.

\bibitem[{Said(1977)}]{said1977}
Said, E.~W. 1977.
\newblock Orientalism.
\newblock \emph{The Georgia Review}, 31(1): 162--206.

\bibitem[{Said(2000)}]{said-2000}
Said, E.~W. 2000.
\newblock \emph{Out of Place—A Memoir}.
\newblock United Kingdom: Vintage Books.

\bibitem[{Sambasivan et~al.(2021)Sambasivan, Kapania, Highfill, Akrong, Paritosh, and Aroyo}]{Sambasivan-2021}
Sambasivan, N.; Kapania, S.; Highfill, H.; Akrong, D.; Paritosh, P.; and Aroyo, L.~M. 2021.
\newblock "Everyone wants to do the model work, not the data work": Data Cascades in High-Stakes AI.
\newblock In \emph{Proceedings of the 2021 CHI Conference on Human Factors in Computing Systems}. New York, USA: ACM.

\bibitem[{Samuels(2020)}]{Samuels-2020}
Samuels, E. 2020.
\newblock How misinformation on WhatsApp led to a mob killing in India.

\bibitem[{Sap et~al.(2019)Sap, Card, Gabriel, Choi, and Smith}]{sap-2019}
Sap, M.; Card, D.; Gabriel, S.; Choi, Y.; and Smith, N.~A. 2019.
\newblock The Risk of Racial Bias in Hate Speech Detection.
\newblock In \emph{Proceedings of the 57th Annual Meeting of the Association for Computational Linguistics}, 1668--1678. Florence, Italy: Association for Computational Linguistics.

\bibitem[{Scarcella(2024)}]{Scarcella-2024}
Scarcella, M. 2024.
\newblock Elon Musk's X wins appeal to block part of California content moderation law.

\bibitem[{Scheuerman and Brubaker(2024)}]{Scheuerman-2024}
Scheuerman, M.~K.; and Brubaker, J.~R. 2024.
\newblock Products of Positionality: How Tech Workers Shape Identity Concepts in Computer Vision.
\newblock In \emph{Proceedings of the 2024 CHI Conference on Human Factors in Computing Systems}, CHI '24. New York, NY, USA: ACM.

\bibitem[{Sch{\"o}pf(2020)}]{schopf-2020}
Sch{\"o}pf, C.~M. 2020.
\newblock The Coloniality of Global Knowledge Production: Theorizing the Mechanisms of Academic Dependency.
\newblock \emph{Social Transformations: Journal of the Global South}, 8(2): 5--46.

\bibitem[{Schwartz(2022)}]{schwartz-2022}
Schwartz, L. 2022.
\newblock {P}rimum {N}on {N}ocere: {B}efore working with {I}ndigenous data, the {ACL} must confront ongoing colonialism.
\newblock In \emph{Proceedings of the 60th Annual Meeting of the Association for Computational Linguistics (Volume 2: Short Papers)}, 724--731. Dublin, Ireland: Association for Computational Linguistics.

\bibitem[{Shahid and Vashistha(2023)}]{Shahid-2023}
Shahid, F.; and Vashistha, A. 2023.
\newblock Decolonizing Content Moderation: Does Uniform Global Community Standard Resemble Utopian Equality or Western Power Hegemony?
\newblock In \emph{Proceedings of the 2023 CHI Conference on Human Factors in Computing Systems}. New York, USA: ACM.

\bibitem[{Siapera(2022)}]{siapera-2022}
Siapera, E. 2022.
\newblock AI Content Moderation, Racism and (de) Coloniality.
\newblock \emph{International Journal of Bullying Prevention}, 4(1): 55--65.

\bibitem[{Sreelekha, Bhattacharyya, and Malathi(2018)}]{sreelekha-2018}
Sreelekha, S.; Bhattacharyya, P.; and Malathi, D. 2018.
\newblock Statistical vs. rule-based machine translation: A comparative study on indian languages.
\newblock In \emph{International Conference on Intelligent Computing and Applications: ICICA 2016}, 663--675. Australia: Springer.

\bibitem[{Stokel-Walker(2024)}]{Walker-2024}
Stokel-Walker, C. 2024.
\newblock Under Elon Musk, X is denying API access to academics who study misinformation.

\bibitem[{stream(2025{\natexlab{a}})}]{audio-2025}
stream. 2025{\natexlab{a}}.
\newblock Audio Moderation.

\bibitem[{stream(2025{\natexlab{b}})}]{video-2025}
stream. 2025{\natexlab{b}}.
\newblock Video Moderation.

\bibitem[{Thakur(2025)}]{Thakur-2025}
Thakur, D. 2025.
\newblock Moderating Quechua Content on Social Media.
\newblock Technical report, Center for Democracy and Technology.

\bibitem[{Thiong'o(1986)}]{Thiong-1986}
Thiong'o, N.~u.~i.~w. 1986.
\newblock \emph{Decolonising the Mind: The Politics of Language in African Literature}.
\newblock East Africa: EAEP.

\bibitem[{TikTok(2024)}]{tiktok}
TikTok. 2024.
\newblock Supporting independent research.

\bibitem[{Udupa, Maronikolakis, and Wisiorek(2023)}]{udupa-2023}
Udupa, S.; Maronikolakis, A.; and Wisiorek, A. 2023.
\newblock Ethical scaling for content moderation: Extreme speech and the (in) significance of artificial intelligence.
\newblock \emph{Big Data \& Society}, 10(1): 1--15.

\bibitem[{Union(2025)}]{union-2025}
Union, U.~G. 2025.
\newblock Content moderators launch first-ever global alliance, demand safe working conditions and accountability from tech giants.

\bibitem[{van Esch et~al.(2019)van Esch, Sarbar, Lucassen, O'Brien, Breiner, Prasad, Crew, Nguyen, and Beaufays}]{van-2019}
van Esch, D.; Sarbar, E.; Lucassen, T.; O'Brien, J.; Breiner, T.; Prasad, M.; Crew, E.; Nguyen, C.; and Beaufays, F. 2019.
\newblock Writing across the world's languages: Deep internationalization for Gboard, the Google keyboard.

\bibitem[{Varshney(2024)}]{varshney2024decolonial}
Varshney, K.~R. 2024.
\newblock Decolonial AI Alignment: Openness, Vi\'{s}esa-Dharma, and Including Excluded Knowledges.
\newblock In \emph{Proceedings of the AAAI/ACM Conference on AI, Ethics, and Society}, volume~7 of \emph{AIES '24}, 1467--1481. New York, NY, USA: ACM.

\bibitem[{Verran and Christie(2007)}]{verran-2007}
Verran, H.; and Christie, M. 2007.
\newblock Using/designing digital technologies of representation in Aboriginal Australian knowledge practices.
\newblock \emph{Human Technology}, 3(2): 214--227.

\bibitem[{Villenas(1996)}]{villenas1996colonizer}
Villenas, S. 1996.
\newblock The colonizer/colonized Chicana ethnographer: Identity, marginalization, and co-optation in the field.
\newblock \emph{Harvard educational review}, 66(4): 711--732.

\bibitem[{Wei, Zufall, and Jia(2025)}]{Wei-2025}
Wei, J. T.-Z.; Zufall, F.; and Jia, R. 2025.
\newblock Operationalizing Content Moderation "Accuracy" in the Digital Services Act.
\newblock In \emph{Proceedings of the 2024 AAAI/ACM Conference on AI, Ethics, and Society}, AIES '25, 1527–1538. New York, NY, USA: ACM.

\bibitem[{Witness(2022)}]{witness-2022}
Witness, G. 2022.
\newblock Facebook unable to detect hate speech weeks away from tight Kenyan election.

\bibitem[{Wong and Ernst(2021)}]{wong-Ernst-2021}
Wong, J.~C.; and Ernst, J. 2021.
\newblock Facebook knew of Honduran president’s manipulation campaign – and let it continue for 11 months.

\bibitem[{Wong and Harding(2021)}]{wong-harding-2021}
Wong, J.~C.; and Harding, L. 2021.
\newblock `Facebook isn't interested in countries like ours': Azerbaijan troll network returns months after ban.

\bibitem[{Yibeltal and Muia(2023)}]{Yibeltal-2023}
Yibeltal, K.; and Muia, W. 2023.
\newblock Facebook's algorithms `supercharged' hate speech in Ethiopia's Tigray conflict.

\bibitem[{Zevallos and Bel(2023)}]{zevallos-bel-2023-hints}
Zevallos, R.; and Bel, N. 2023.
\newblock Hints on the data for language modeling of synthetic languages with transformers.
\newblock In \emph{Proceedings of the 61st Annual Meeting of the Association for Computational Linguistics (Volume 1: Long Papers)}, 12508--12522. Toronto, Canada: Association for Computational Linguistics.

\bibitem[{Zhong et~al.(2024)Zhong, Yang, Liu, Zhang, Liu, Sun, Pan, Li, Zhou, Jiang et~al.}]{zhong2024}
Zhong, T.; Yang, Z.; Liu, Z.; Zhang, R.; Liu, Y.; Sun, H.; Pan, Y.; Li, Y.; Zhou, Y.; Jiang, H.; et~al. 2024.
\newblock Opportunities and challenges of large language models for low-resource languages in humanities research.

\end{thebibliography}

\end{document}